%% file: main.tex
\def\year{2019}\relax
\documentclass[letterpaper]{article} %DO NOT CHANGE THIS
\usepackage{aaai19}  %Required
\usepackage{times}  %Required
\usepackage{helvet}  %Required
\usepackage{courier}  %Required
\usepackage{url}  %Required
\usepackage{graphicx}  %Required
\usepackage{times}
\usepackage{caption}
\usepackage{multirow}
\usepackage{ifthen}
\usepackage{enumitem}
\usepackage{todonotes}
\usepackage{changepage}
\usepackage{amsmath}
\usepackage{caption}
\usepackage{multirow}
\usepackage{ifthen}
\usepackage{enumitem}
\usepackage{float}
\usepackage{placeins}
\usepackage{amsfonts}
\usepackage{mathtools}
\usepackage{booktabs}
\usepackage{subcaption}

\DeclarePairedDelimiter\floor{\lfloor}{\rfloor}
\usepackage[english]{babel}
\usepackage[utf8]{inputenc}
\usepackage{algorithm}
\usepackage[noend]{algpseudocode}
\usepackage{wrapfig}
\algnewcommand\algorithmicforeach{\textbf{for each}}
\algdef{S}[FOR]{ForEach}[1]{\algorithmicforeach\ #1\ \algorithmicdo}
\begin{document}
%\title{On Low Memory Footprint Deep NLP Models \\
%	Preserving Accuracy through Spectral Compression Aware Training}   
%\title{Compression of Text Classification Models using Low Rank Matrix Factorization }
%\title{Lossless Embedding  using Low Rank Matrix Factorization }
%\title{Using Low Rank Matrix Factorization for NLP Model Compression }    
\title{Online Embedding Compression for Text Classification using\\
	  Low Rank Matrix Factorization}
  \author{Anish Acharya \thanks{Corresponding Author} \textsuperscript{1,3}, Rahul Goel \textsuperscript{1}, Angeliki Metallinou \textsuperscript{1}, Inderjit Dhillon \textsuperscript{2,3} \\ \textsuperscript{1} Amazon Alexa AI , \textsuperscript{2} Amazon Search Technologies, \textsuperscript{3} University of Texas at Austin\\  {achanish, goerahul, ametalli \}@amazon.com}, isd@a9.com }
\maketitle
\begin{abstract}
Deep learning models have become state of the art for natural language processing (NLP) tasks, however deploying these models in production system poses significant memory constraints. Existing compression methods are either lossy or introduce significant latency. We propose a compression method that leverages low rank matrix factorization during training,to compress the word embedding layer which represents the size bottleneck for most NLP models. Our models are trained, compressed and then further re-trained on the downstream task to recover accuracy while maintaining the reduced size. Empirically, we show that the proposed method can achieve 90\% compression with minimal impact in accuracy for sentence classification tasks, and outperforms alternative methods like fixed-point quantization or offline word embedding compression. We also analyze the inference time and storage space for our method through FLOP calculations, showing that we can compress DNN models by a configurable ratio and regain accuracy loss without introducing additional latency compared to fixed point quantization. Finally, we introduce a novel learning rate schedule, the Cyclically Annealed Learning Rate (CALR), which we empirically demonstrate to outperform other popular adaptive learning rate algorithms on a sentence classification benchmark.

\end{abstract}
\input{intro}

\input{related}
\input{method}

\input{analysis}
\input{result}

\vspace{-1mm}
\input{conclusion}

\vspace{-3mm}
\bibliography{compression}
\bibliographystyle{aaai}
\end{document}

%% file: intro.tex
\section{Introduction}\label{intro}
%% background %% 
%In the past decade deep neural nets have achieved state of the art performance on a range of speech, vision and NLP tasks. 

Deep learning has achieved great success in various NLP tasks such as sequence tagging~\cite{chung_rnn},~\cite{ma_hovy_2016} and sentence classification~\cite{liu_multi_timescale,conv_text_kim}.
%% The NLP literature applies and combines common
%% neural models including LSTMs~\cite{hochreiter_lstm}, CNNs~\cite{lecun98},
%% DANs~\cite{iyyer2015deep} etc.
While traditional machine learning approaches
extract hand-designed and task-specific features, which are fed into a shallow
model, neural models pass the input through several feedforward, recurrent or
convolutional layers of feature extraction that are trained to learn
automatic feature representations. However, deep neural models often have
large memory-footprint and this poses significant deployment challenges for
real-time systems that typically have memory and computing power constraints. For
example, mobile devices tend to be limited by their CPU speed, memory and
battery life, which poses significant size constraints for models embedded on
such devices. Similarly, models deployed on servers need to serve millions of
requests per day, therefore compressing them would result in memory and
inference cost savings. 

For NLP specific tasks, the word embedding matrix often
accounts for most of the network size. The embedding matrix is typically
initialized with pretrained word embeddings like Word2Vec, ~\cite{mikolov2013a}
FastText~\cite{bojanowski2016enriching} or Glove ~\cite{pennington2014glove} and then fine-tuned on the
downstream tasks, including tagging, classification and others. Typically, word
embedding vectors are 300-dimensional and vocabulary sizes for practical
applications could be up to 1 million tokens. This corresponds to up to 2Gbs of
memory and could be prohibitively expensive depending on the application. For
example, to represent a vocabulary of 100K words using 300 dimensional Glove
embeddings the embedding matrix would have to hold 60M parameters. Even for a
simple sentiment analysis model the embedding parameters are 98.8\% of the total
network~\cite{shu2017compressing}.  \\
In this work, we address neural model compression in the context of text
classification by applying low rank matrix factorization on the word embedding layer and then re-training the model in an online fashion. There has been relatively less work in compressing word embedding matrices of deepNLP models. Most of the prior work compress the embedding matrix offline outside the training loop either through hashing or quantization based approaches \cite{joulin2016fasttext,shu2017compressing,raunak2017effective}.
%Those include works by~\citeauthor{shu2017compressing}, who do hashing on the embeddings and more recently~\citeauthor{joulin2016fasttext}, who used product quantization for compression.
%This is different from
%other related works on word embedding size reduction like ~\cite{raunak2017effective,joulin2016fasttext}, which compress the embedding matrix offline outside the training loop. 
Our approach includes starting from a model initialized with large embedding space, performing a low rank projection of the embedding layer using Singular Value Decomposition (SVD) and continuing training to regain any lost accuracy. This enables us to compress a deep NLP model by an arbitrary compression fraction ($p$), which can be
pre-decided based on the downstream application constraints and accuracy-memory
trade-offs. Standard quantization techniques do not offer this flexibility, as
they typically allow compression fractions of $1/2$ or $1/4$ (corresponding to
16-bit or 8-bit quantization from 32-bits typically). \\
We evaluate our method on text classification tasks, both on the benchmark SST2 dataset and on a proprietary dataset from a commercial artificial agent. We show that our method significantly reduces the model size (up to 90\%) with minimal accuracy impact (less than 2\% relative), and outperforms popular compression techniques including quantization~\cite{hubara2016quantized} and offline word embedding size reduction~\cite{raunak2017effective}.\\
A second contribution of this paper is the introduction of a novel learning rate schedule,we call Cyclically Annealed Learning Rate (CALR), which extends previous works on cyclic learning rate~\cite{smith2017cyclical} and random hyperparameter search~\cite{bergstra2012random}. Our experiments on SST2 demonstrate that for both DAN~\cite{iyyer2015deep} and LSTM~\cite{hochreiter_lstm} models CALR outperforms the current state-of-the-art results that are typically trained using popular adaptive learning like AdaGrad. \\
% schedule ourperforms popular adaptive learning rate algorithms like Adagrad on the SST2 benchmark dataset.
Overall, our contributions are: %\vspace{-3.7 pt}
\begin{itemize}[noitemsep, topsep=0pt ]
%\item We propose a method to compress deep NLP models that can reduce the memory footprint through spectrally projecting embedding space into a low dimensional space during training.
\item We propose a compression method for deep NLP models that reduces the memory
footprint through low rank matrix factorization of the embedding layer and regains accuracy through further finetuning.
\item We empirically show that our method outperforms popular baselines like fixed-point quantization and offline embedding compression for sentence classification.
\item We provide an analysis of inference time for our method, showing that we can compress models by an arbitrary configurable ratio, without introducing additional latency compared to quantization methods.
\item We introduce CALR, a novel learning rate scheduling algorithm for gradient descent based optimization and show that it outperforms other popular adaptive learning rate algorithms on sentence classification.
\end{itemize}

%% file: related.tex
\section{Related Work} 
\label{sec:related}
In recent literature, training overcomplete respresentations is often advocated as overcomplete bases can transform local minima into saddle points \cite{dauphin2014identifying} or help discover robust solutions \cite{lewicki2000learning}. This indicates that significant model compression is often possible without sacrificing accuracy and that the model's accuracy does not rely
on precise weight values~\cite{keskar2016large}. Most of the recent work on model compression exploits this inherent sparsity of overcomplete represantations. These works include low precision computations~\cite{anwar2015fixed,courbariaux2014low} and quantization of model weights~\cite{han2015deep,zhou2017incremental}. There are also methods which
prune the network by dropping connections with low weights ~\cite{wen2016learning,see2016compression} or use sparse encoding ~\cite{han2015learning}. These methods in practice often suffer from quantization loss especially if the network is deep due to a large number of low-precision multiplications during forward pass. Quantization loss is hard to recover through further training due to the non-trivial nature of backpropagation in low precision~\cite{lin2015neural}. There has been some work on compression aware training~\cite{polino2018model} via model distillation~\cite{hinton2015distilling} that addresses this gap in accuracy. However, both quantized distillation and differentiable quantization methods introduce additional latency due to their slow training process that requires careful tuning, and may not be a good fit for practical systems where a compressed model needs to be rapidly tuned and deployed in production. \\
%We explore a different perspective to approach the compression problem. We argue that since the parameter space of a DNN is often overcomplete, the representation of an input is not an unique combination of basis vectors (Lewicki and Sejnowski 2000). This means that it is possible to find a lower dimensional parameter space which can represent the model with minimal information loss. %Extending this intuition, we use low rank matrix factorization techniques for compressing large weight matrices in deep
%models, focusing on the word embedding matrix which is the size bottleneck for many NLP tasks
 %Rahul:this feels incomplete
Low-rank matrix factorization (LMF) is a very old dimensionality reduction technique widely used in the matrix completion
literature. For example, see \cite{recht2013parallel} and references therein. 
However, there has been  relatively limited work on applying LMF to deep neural models. \cite{sainath2013low,lu2016learning} used low rank matrix factorization for neural speech models.While~\cite{sainath2013low}
reduces parameters of DNN before training,~\cite{xue2013restructuring}
restructures the network using SVD introducing bottleneck layers in the weight
matrices. However, for typical NLP tasks, introducing bottleneck layers between the deep layers of the model does not significantly decrease model size since the large majority of the parameters in NLP models come from the input word embedding matrix.
Building upon this prior work, we apply LMF based ideas to reduce the embedding space of NLP models. Instead of exploiting the sparsity of parameter space we argue that for overcomplete representation the input embedding matrix is not a unique combination of basis vectors~\cite{lewicki2000learning}. Thus, it is possible to find a lower dimensional parameter space which can represent the model inputs with minimal information loss. Based on this intuition, we apply LMF techniques for compressing the word embedding matrix which is the size bottleneck for many NLP tasks.

%% file: method.tex
\section{Methodology}
\label{sec:method}
\subsection{Low-rank Matrix Factorization (LMF) for Compressing Neural Models}
\label{sec:method_svd}
Low-rank Matrix Factorization (LMF) exploits latent structure in the data to obtain a
compressed representation of a matrix. It does so by factorization of the
original matrix into low-rank matrices. For a full rank matrix $W
\in R^{m\times n}$ of rank $r$, there always exists a factorization $W=W_a
\times W_b$ where $W_a \in \mathbb{R}^{m \times r}$ and $W_b \in \mathbb{R}^{r \times n}$. Singular Value Decomposition (SVD) \cite{golub1970singular} achieves this factorization as follows:
\begin{equation}\label{eq1}
W_{m\times n} = U_{m\times m}\Sigma_{m\times n}V^{t}_{n\times n} 
\end{equation}
where $\Sigma_{m\times n}$ is a diagonal rectangular matrix of singular values of descending
magnitude.  We can compress the matrix by choosing the $k$ largest singular
values, with $k<m$ and $k<n$. For low rank matrices, the discarded singular values will be zero or small, therefore the following will approximately hold:
\begin{equation}\label{eq2}
W_{m\times n} \approx U_{m\times k}\Sigma_{k\times k}V^{t}_{k\times n} \newline
%= U_{m\times k}N_{k\times n}
\end{equation}

Therefore, the original matrix $W$ is compressed into two matrices:
\begin{equation}\label{eq3}
W_a = U \in \mathbb{R}^{m \times k} \newline
\end{equation}
\begin{equation}\label{eq4}
W_b = \Sigma_{k\times k}V^{t}_{k\times n} \in \mathbb{R}^{k \times n}
\end{equation}

The number of parameters are reduced from $m \times n$ to $k\times(m +
n)$. Therefore, to achieve a $p$ fraction parameter reduction, $k$ should be
selected as:
\begin{align}\label{eq5}
(p \times m \times n) =  k \times (m + n) \notag \\
%\end{equation}
%\begin{equation}\label{eq6}
k = \floor* {\frac{pmn}{m + n}} ;  k \in \mathbb{Z}^{+} 
\end{align}

\begin{figure}[th]
	\centering
	\includegraphics[width=0.35\textwidth]{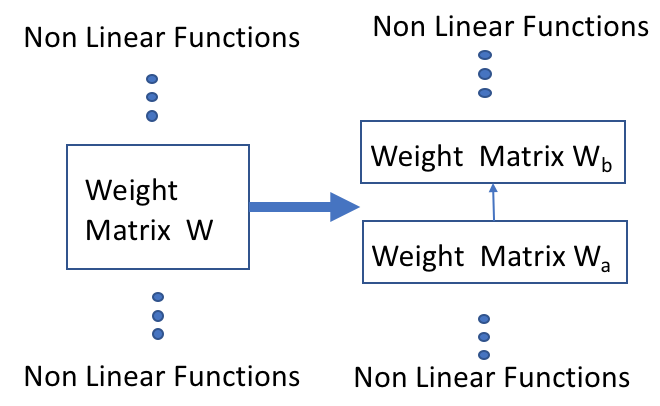}
	\caption{Replacing one neural network matrix with two low rank matrices}
	\label{fig:decomposition}
\end{figure}

The value $k$ can be varied  depending on the desired compression fraction $p$ as long as $k \geq 1$. In practice, being able to decide an arbitrary compression fraction $p$ is an attractive property for model compression, since it allows choosing compression rates based on the accuracy-memory trade-offs of a downstream application. 
%Standard quantization techniques do not offer this flexibility, as they typically allow compression fractions of $1/2$ or $1/4$ (corresponding to 16-bit or 8-bit quantization from 32-bits typically).
The low rank matrix factorization operation is illustrated in
Figure~\ref{fig:decomposition}, where a single neural network matrix (layer) is
replaced by two low rank matrices (layers).
\vspace{-8 pt}
\subsubsection{Compressing the Word Embedding Matrix }
%% Needs Major Revision 
For NLP tasks, typically the first layer of the neural model consists of an
embedding lookup layer that maps the words into real-valued vectors for
processing by subsequent layers. The words are indices $i$ of a word dictionary
of size $D$ while the word embeddings are vectors of size $d$. This corresponds
to a lookup table $E(i) = W_i$ where $W \in \mathbb{R}^{d \times |D|}$. The word
embeddings are often trained offline on a much larger corpus, using algorithms
like Word2Vec, GloVe etc, and are then fine-tuned during training on the downstream NLP task.  Our method 
decomposes the embedding layer in an online fashion during training using eq.(\ref{eq5}).
$W_a$ becomes the new embedding layer and $W_b$ becomes the next layer. Continuing backpropagation over this new network struture finetunes the low rank embedding space and regains any accuracy loss within a few epochs.
\subsection{Cyclically Annealed Learning Rate }
\label{subsec:method_training_scheme}
\begin{figure}[!htb]
 	%\minipage{0.2365\textwidth}	
 	\begin{subfigure}{.23\textwidth}
 		\centering
 		\includegraphics[width=\linewidth]{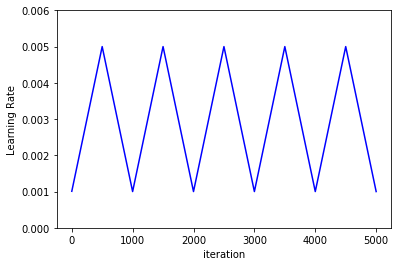}\caption{CLR}	\label{clr} 
 	\end{subfigure}
 	%\caption{}
 	%\endminipage\hfill
 	%\minipage{0.2365\textwidth}
 	\begin{subfigure}{.23\textwidth}	
 		\centering
 		\includegraphics[width=\linewidth]{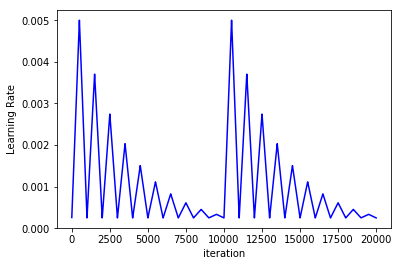}\caption{CALR}	\label{modified_clr} 
 		%\caption{}\label{2b}
 		%\endminipage\hfill
 	\end{subfigure}
 	\caption{Comparison of CLR and CALR update Policy }
 \end{figure}

Due to the non-convex nature of the optimization surface of DNNs, gradient based algorithms like stochastic gradient descent (SGD) are prone to getting trapped in suboptimal local minima or saddle points. Adaptive learning rates like Adam~\cite{kingma2014adam} and Adagrad ~\cite{duchi2011adaptive} try to solve this problem by tuning learning rate based on the magnitude of gradients. While in most cases they find a reasonably good solution, they usually can't explore the entire gradient landscape. \cite{bergstra2012random,loshchilov2016sgdr} indicates periodic random initialization or warm restarts often helps in better exploration of the landscape while \cite{smith2017cyclical} showed that letting the learning rate(LR) vary cyclicaly (Cyclic learning rate(CLR)) helps in faster convergence.  
In this work we further refine CLR and propose a simulated annealing inspired LR update policy we call Cyclically Annealed Learning
Rate~(CALR). CALR is described in algorithm (\ref{alg:Exp_clr}) and Fig. \ref{modified_clr}. In addition to varying LR in a CLR style triangular windows(Fig.~\ref{clr}), CALR expontentially decays the upper bound $LR_{UB}$ of the triangular window from its initial value. When the learning rate decreases to a lower bound $LR_{LB}$, we increase the upper bound to its initial value $LR_{UB}$ and continue with the LR updates. This exponential decay ensures slow steps on the optimization
landscape. Intuitively, the proposed warm restarts, motivated by simulated annealing, make
CALR more likely to escape from local minima or saddle points and enables better
exploration of the parameter space.  CALR should have similar but less
aggressive effect as random initialization or warm restarts of LR. While exponentially decayed small
cyclic windows let CALR carefully explore points local to the current solution,
cyclic temperature increase helps it jump out and explore regions around other
nearby local minima. We verified this intuition empirically for our
classification experiments~(Table~\ref{results:uncompressed}, Secton \ref{sec:experiments_uncompresed}), where CALR was
able to achieve further improvement compared to CLR and Adagrad.

\begin{algorithm}
	\caption{Cyclically Annealed Learning Rate Schedule}\label{alg:Exp_clr}
	\begin{algorithmic}[1]
		\Procedure{CALR}{Iteration, Step Size, $LR_{LB}$, $LR_{UB}$}
		\State $LR_{UB} \gets LR_{UB} (init) $
		\State $LR_{LB} \gets LR_{LB} (init) $
		%\State $LR_{LB} \gets LR_{UB} (init) \times \exp(Decay)^{ResetStep}$
		%\Comment{This ensures if $LR_{UB}$ is decayed exponentially by the Decay Rate it will become $\leq LR_{LB}$ after ResetStepSize Epochs}
		\ForEach{EPOCH}
		\State $LR_{UB} \gets LR_{UB}\times \exp(Decay)$
		\If{$LR_{UB} \leq LR_{LB}$}
		\State $LR_{UB} \gets LR_{UB} (init)$
		%\State $LR_{LB} \gets LR_{UB} (init) \times \exp(Decay)^{ResetStep}$
		\EndIf
		\ForEach{Training BATCH}
		\State LR $\gets$ CLR(Iteration,StepSize,$LR_{LB}, LR_{UB}$)
		\EndFor
		\EndFor
		\Return $LR$
		\EndProcedure
	\end{algorithmic}
	\begin{algorithmic}[1]
		\Procedure{CLR}{Iteration, Step Size, $LR_{LB}$, $LR_{UB}$}
		\newline
		\Comment{this procedure implements a triangular window of width StepSize and height $LR_{UB} - LR_{LB}$}
		\State $Bump\gets \frac{LR_{UB} - LR_{LB}}{Step Size}$
		\State $Cycle \gets (Iteration) \bmod (2 \times StepSize)$
		\If{$Cycle < StepSize$}
		\State $LR \gets LR_{LB} + (Cycle \times Bump)$
		\Else 
		\State $LR \gets LR_{UB} - (Cycle - StepSize)\times Bump$
		\EndIf
		\Return $LR$
		\EndProcedure
	\end{algorithmic}
\end{algorithm} 

\subsection{Baselines}
\label{subsec:method_baselines}
We compare our proposed method with two commonly baselines for neural model compression.
%% We compare our proposed method with two baselines that has been used in prior state-of-art method for compressing embedding space of deep NLP models.
\subsubsection{Fixed Point Weight Quantization (Baseline 1)}
We compare our approach with widely used post-training fixed point quantization
\cite{bengio2013estimating,gong2014compressing} method where the weights are
cast into a lower precision. This requires less memory to store them and doing low
precision multiplications during forward propagation improves inference latency.
%Several model weight quantization algorithms have been proposed in the
%literature, here we use tensorflow's ~\cite{abadi2016tensorflow} native
%quantization as our second baseline.
%[@Anish describe the details of the quantization baseline]
\subsubsection{Offline Embedding Compression (Baseline 2)}
In this set of baseline experiments we compress the embedding matrix in a
similar way using SVD but we do it as a preprocessing step similar to
~\cite{raunak2017effective,joulin2016fasttext}. For example, if we have a 300
dimensional embedding and we want a 90\% parameter reduction we project it onto
a 30 dimensional embedding space and use this low dimensional embedding to train
the NLP model.

%% file: analysis.tex
\section{Analysis}
\label{sec: analysis}
In this section we analyze space and latency trade-off in compressing deep
models. We compare the inference time and storage space of floating point operations(FLOPs) between our method and Baseline1(sec.\ref{subsec:method_baselines})  on a 1-layer Neural Network(NN) whose forward pass on input X can be represented as:
\begin{align}\label{eq:7}
f(X) = \sigma(XW) 
\end{align}
where $X \in \mathbb{R}^{1 \times m} \notag $ and $W \in \mathbb{R}^{m \times
  n}$ and full rank. Our method reduces W into two low rank matrices $W_a$
and $W_b$ converting (\ref{eq:7}) into two layer NN represented as:
\begin{align}
\label{eq:8}
f(X) = \sigma ((X\times W_a)\times W_b)
\end{align}
where $W_a \in \mathbb{R}^{m \times k}$ and $W_b \in \mathbb{R}^{k \times n}$
are constructed as described in eq.(\ref{eq3}) and eq.(\ref{eq4}) choosing k as
in eq.(\ref{eq5}).
%Unlike our spectral projection approach, low precision compression (quantization) 
Whereas, Baseline1 casts the elements of $W$ to lower
precision without restructuring the network. Thus, the quantized representation
of a 1-layer NN is same as eq.(\ref{eq:7}) where $W_i$ is cast to a lower bit
precision and eq. (\ref{eq:8}) is the corresponding representation in our approach
where $W_i$ remains in its original precision.
\subsection{Space Complexity Analysis}
\label{sec:analysis_space}
On Baseline1 each weight variable uses less number of bits for
storage. 
%Let's assume storing a weight variable in full precision requires $B_S$ bits and in lower precision requires $B_Q$ bits. 
Quantizing the network weights into
low precision, we achieve space reduction by a fraction $\frac{B_Q}{B_S}$ where $B_S$ bits are required to store a full precision weight and $B_Q$ bits are needed to store low precision weight. In contrast, our approach described in section \ref{sec:method_svd}, achieves more
space reduction as long as we choose $k$ (from eq (\ref{eq5})) such that:
\begin{equation} \label{eq:9}
p < \frac{B_Q}{B_S}
\end{equation}
%In practice, most compression methods use 16-bit or 8-bit quantization which achieves $1/2$ or $1/4$ compression respectively, assuming original precision of 32-bits (which is used in popular toolboxes like tensorflow). Therefore, for our proposed method choosing $k$ such as $p<1/4$ would achieve more size compression than typical quantization methods in the literature.
\subsection{Time Complexity Analysis}
\label{sec:analysis_time}
Total number of FLOPs in multiplying two matrices~\cite{trefethen1997numerical}
is given by: $FLOP(A \times B) = (2b - 1)ac \sim O(abc) $
%\begin{align} \label{eq:10} %\end{align}
where $A \in \mathbb{R}^{a\times b}$, $B \in \mathbb{R}^{b\times c}$ and $AB \in
\mathbb{R}^{a\times c}$.  Assuming one FLOP in a low precision matrix
operations takes $T_Q$ time and on a full precision matrix it takes $T_S$
time,  for the model structure given by eq. (\ref{eq:7}), we have: $a=1, b=m,
c=n$. Hence, for the forward pass on a single data vector $X \in \mathbb{R}^{1
  \times m}$ a quantized model uses $F_Q$ flops where:
\begin{equation}\label{eq:11}
F_Q = (2m -1)n 
\end{equation}
Our algorithm restructures the DNN given in eq. (\ref{eq:7}) into eq. (\ref{eq:8})
yielding $F_S = F_{S1} + F_{S2}$ flops. where:
\begin{align}
\label{eq:11}
F_{S1} = (2m-1)k, \  F_{S2} = (2k-1)n  \notag \\
F_S = 2(m+n)k-(n+k)
\end{align}
The number of FLOPs in one forward pass our method would be less than that of in
Baseline1 if $F_Q < F_S$ or equivalently:
\begin{align}
\label{eq:13}
(2m -1)n &> 2(m+n)k -(n+k) \notag \\
%2mn &> 2(m + n)k - k \notag \\
k &< \frac{2mn}{2(m+n) - 1}
\end{align}
Under the reasonable assumption that $2(m+n) \gg 1$ or equivalently $2(m+n) - 1 \approx 2(m+n)$,
eq. (\ref{eq:13}) becomes:
\begin{align}\label{eq:14}
k &< \frac{mn}{(m+n)}
\end{align}
Eq. (\ref{eq:14}) will hold as long as $k$ in our method is chosen according to
eq.(\ref{eq5}) (since the fraction $p$ is by definition $p<1$)
However, guaranteeing less FLOPs does not guarantee faster inference since
quantized model requires less time per FLOP. 
%Assuming that $T_Q$ is the time per
%FLOP for the quantized model and $T_S$ for 
Our method guarantees
faster inference if the following condition holds:
\begin{align} \label{eq:15}
F_Q T_Q &> F_S T_S \notag\\
(2m -1)n T_Q &> [(2m -1)k + (2k -1)n] T_S \notag \\
\frac{(2m -1)n}{[(2m -1)k + (2k -1)n]} &> \frac{T_S}{T_Q} 
\end{align}
where $T_S$ and $T_Q$ are time taken per FLOP for our method and Baseline1 respectivly.
Again, we can make the reasonable assumptions that $2m \gg 1$ and $2k \gg 1 $,
therefore $2m - 1 \approx 2m$ etc. Then eq. (\ref{eq:15}) becomes:
\begin{align} \label{eq:16}
\frac{mn}{(m + n) k} &> \frac{T_S}{T_Q}
\end{align}
Plugging $k$ from eq.(\ref{eq5}) we have: 
\begin{align}\label{eq:17}
p &< \frac{T_Q}{T_S} 
\end{align}
If we assume that the time per FLOP is proportional to the number of bits $B$
for storing a weight variable, e.g., $T_Q \sim B_Q$ and $T_S \sim B_S$ then
selecting $p$ such that $p < \frac{B_Q}{B_S} $ will ensure that
eq. (\ref{eq:17}) holds. In most practical scenarios time saving is sublinear
in space saving which means eq.(\ref{eq:9}) is necessary and sufficient to
ensure our method has faster inference time than fixed point quantization(Baseline1).

%% file: result.tex
\section{Experiments and Results}
\label{sec:experiments}
Our experiments are focused on sentence classification task. %where the task is to assign a label to a given sentence. 
%\subsection{Experimental Setup}
%These numbers might be different in other frameworks but should have similar trends.
\subsection{Datasets:} %\\
For all the experiments reported we have used the following two sentence classification datasets:    \\
%\subsubsection{Movie Review Dataset (MR)}
%The Movie Review dataset~\cite{maas2011learning} is a common benchmark classification dataset that contains movie reviews, that should be classified as either positive or negative. On an average the reviews are about 200 words long. It has balanced train and test sets containing 25k examples each.
\textbf{Stanford Sentiment Treebank (SST2)}
SST2 ~\cite{socher2013recursive} is a common text classification benchmark dataset that contains movie reviews, to be classified according to their sentiment. On an average the reviews are about 200 words long. The dataset contains standard train, dev, test splits and binary sentiment labels. \\
\textbf{Books Intent Classification Dataset}
We use a proprietary dataset of around 60k annotated utterances of users interacting with a popular digital assistant about books related functionality. Each utterance is manually labeled with the user intent e.g., search for a book, read a book and others, with the task being to predict the user intent from a set of 20 intents. We choose 50k utterances as our training data, 5k as test data and 5k as dev data.%\\ \\
\subsection{Models:}%\\
We evaluate our compression strategy on two types of widely used neural models: \\ %which are commonly used for classification tasks:\\
\textbf {Long Short Term Memory(LSTM)}
LSTMs, are powerful and popular sequential models
for classification tasks. The LSTM units are recurrent inn nature, designed to
handle long term dependencies through the use of input, output
and forget gates and a memory cell. For input $X=x_1,
\cdots x_T $ and corresponding word embedding input vectors $e_i, i=1, \cdots
T$, LSTM computes a representation $r_t$ at each word $t$, which is denoted as
$r_t = \phi ( e_t ,r_{t-1} )$. Here, for sentence classification, we used the representation obtained from the final timestep of the LSTM and
pass it through a softmax for classification:
%\begin{gather}
$r^{sent} = r^{forw}_{T},$ and $ \hat{S}=softmax( W_{s} r^{sent} + b_{s}) $
%\end{gather}
where $r^{sent}$ is the sentence representation, and $\hat{S}$ is the label
predicted. In our experiments with LSTM, we used
300-dimensional Glove embeddings to initialize the embedding layer for all the methods, e.g., baselines 1, 2 and our proposed compression method. Our vocabulary size $V$ contains only the words that appear in the training data, and we compress the input word embedding matrix $W$ (original size $V \times 300$).\\
\textbf{Deep Averaging Network (DAN)} DAN is a
bag-of-words neural model that averages the word embeddings in each input
utterance to create a sentence representation $r^{sent}$ that is passed
through a series of fully connected layers, and fed into a softmax layer for
classification. Assume an input sentence $X$ of length $T$ and
corresponding word embeddings $e_i$, then the sentence representation is: $r^{sent} = \frac{1}{T} \sum_{i=1}^T { e_i }$
%\begin{gather}
%r^{sent} = \frac{1}{T} \sum_{i=1}^T { e_i }
%\end{gather}
%DAN has been proven a state-of-the-art model for simple classification tasks
%such as the Factoid Question Answering task~\cite{iyyer2015deep}. 
In our setup, we have two fully connected layers after the sentence representation $r^{sent}$,
of sizes 1024 and 512 respectively. For the DAN experiments we used the DAN-RAND
~\cite{iyyer2015deep} variant where we take a randomly initialized 300
dimensional embedding. To make the comparison fair, all experiments with DAN models, including quantization (baseline 1), the offline compression method (baseline 2) and the proposed compression method are done using randomly initialized word embedding matrices. For example, our baseline 2 is to use appropriate low-rank randomly initialized embedding matrix. %\\ \\
\subsection{Experimental Setup}%\\
Our experiments are performed on 1 Tesla K80 GPU, using SGD optimizer with CALR policy (Sec.\ref{subsec:method_training_scheme}), initial learning rate upper bound of 0.001. We use dropout of 0.4 between layers and L2 regularization with weight of 0.005. Our metrics include accuracy and model size. Accuracy is simply the fraction of utterances with correctly predicted label. The model size is calculated as sum of (.data + .index + .meta) files stored by tensorflow. 
%Setting up the experiments with randomly initialized embeddings for DAN, as well as GloVe
%initialized ensures that all the gains we report is not due to low dimensional
%latent structure of the initialization embedding matrix but are attributed to
%our proposed approach.
\begin{table}[b]
  \centering
  \scalebox{0.9}{
	\begin{tabular}{lllll} \hline
		Model & LR Schedule & Acc.\\ \hline
		\multicolumn{3}{c}{Uncompressed LSTM Model}\\ \hline
		\citeauthor{tai2015improved}&    AdaGrad       & 84.90  \\
		\citeauthor{smith2017cyclical}& CLR & 84.71\\
		Ours & CALR & \textbf{86.03}\\\hline
		\multicolumn{3}{c}{Uncompressed DAN-RAND Model}\\ \hline
		\citeauthor{iyyer2015deep} &    AdaGrad       & 83.20  \\
		\citeauthor{smith2017cyclical} & CLR & 83.84\\
		Ours & CALR & \textbf{84.61}		
	\end{tabular} }
	\caption{\small Comparison of initial Uncompressed Model for different Learning Rate Schedules on SST2}
	\label{results:uncompressed}	
\end{table}

\subsection{Results on Uncompressed Models}
\label{sec:experiments_uncompresed}
To effectively train our uncompressed benchmark networks we experimented with
different adaptive learning rate schedules. Table~\ref{results:uncompressed}
emperically demonstrates the effectiveness of our proposed learning rate update
policy (CALR) on the SST2 test set. For both the
models CALR outperforms the corresponding state-of-the-art sentence
classification accuracy reported in the literature. For LSTM we improve the
accuracy from 84.9\%~\cite{socher2013recursive} to 86.03\% (1.33\% relative
improvement) whereas on DAN-RAND we are able to improve from 83.2\%
~\cite{iyyer2015deep} to 84.61\% (1.7\% relative improvement). We also report
results on training the network with CLR without our cyclic annealing. For both
DAN-RAND and LSTM, CLR achieves performance that is similar to the previously reported state-of-the-art. The proposed CALR outperforms CLR, which supports the effectiveness of the proposed CALR policy.
%which further strenghthens the effectiveness of CALR policy.

\begin{table}[t]
  \centering
  \scalebox{0.9}{
        \begin{tabular}{lllll} \hline
Model               & R(\%) & Size(MB)    &  Acc. & Acc.\\ \hline
\multicolumn{5}{c}{Uncompressed LSTM  model}\\ \hline
LSTM          & 0             & \bf{53.25} & \bf{86.03}            & -     \\
\hline
\multicolumn{5}{c}{Quantized model (Baseline 1) }\\ \hline
16 bit  & 50            & 30.16 & 85.08              & -        \\
8 bit & 75            & 18.21 & 85.01              & -       \\ \hline 
\multicolumn{5}{c}{SVD: Proposed vs Offline Compression (Baseline 2)}\\\hline
                &               &          &   Proposed   &   Baseline2  \\ 
LSTM            & 10            & 48.66 & 85.72              & 85.45    \\
            & 30            & 38.52 & 85.68              & 84.95     \\
          & 50            & 28.45 & 85.67              & 84.24     \\
          & 70            & 18.38 & 85.45              & 83.09     \\
         & 90            & \bf{6.94}  & \bf{85.11}              & 82.54      \\ \hline\hline
\multicolumn{5}{c}{Uncompressed DAN model}\\\hline
DAN             & 0             & \bf{52.84}  & \bf{84.61}            & -    \\  \hline
\multicolumn{5}{c}{Quantized model (Baseline 1) }\\ \hline
16 bit & 50            & 28.36 & 83.18              & -            \\
8 bit & 75            & 18.13 & 82.94              & -             \\ \hline
\multicolumn{5}{c}{SVD: Proposed vs Offline Compression (Baseline 2)}\\ \hline
             &               &          &   Proposed   &   Baseline2  \\ 
DAN             & 10            & 47.29 & 84.24              & 83.47      \\
                    & 30            & 37.23 & 83.83         & 83.31     \\
                    & 50            & 27.16 & 83.72         & 82.87       \\
                    & 70            & 17.18 & 83.67          & 82.86     \\
                    & 90            & \bf{6.21}  & \bf{83.11}          & 82.59  \\ \hline
\end{tabular}}
	\caption{\small Compression and accuracy results on SST2 dataset. R(\%) refers to percentage of model size reduction, Size is the model size, and Acc is the classification accuracy. All DAN models use the DAN-RAND variant.}
	\label{results:sst}
\end{table}

\begin{table}[t]
  \centering
  \scalebox{0.9}{
	\begin{tabular}{lllll} \hline
		Model               & R(\%) & Size(MB)    &  Acc. & Acc.\\ \hline
		\multicolumn{5}{c}{Uncompressed LSTM model}\\ \hline
		LSTM          & 0             &\bf{29.34} & \bf{91.78}            & -     \\
		\hline
		\multicolumn{5}{c}{Quantized model (Baseline 1) }\\ \hline
		16 bit  & 50            & 17.92  & 90.16             & -        \\
		8 bit & 75            & 10.05  & 89.96              & -       \\ \hline 
		\multicolumn{5}{c}{SVD: Proposed vs Offline Compression (Baseline 2)}\\\hline
		&               &          &   Proposed   &   Baseline2  \\ 
		LSTM            & 10          & 26.76 & 91.71          & 91.07    \\
		& 30            & 21.69 & 91.63            & 90.97     \\
		& 50            & 16.62 & 91.54             & 90.87     \\
		& 70            & 10.31  &91.47              & 90.58     \\
		& 90            & \bf{4.89}  &\bf{90.94}              &89.83      \\ \hline\hline
		\multicolumn{5}{c}{Uncompressed DAN model}\\\hline
		DAN             & 0             & \bf{37.03} & \bf{90.18}            & -    \\  \hline
		\multicolumn{5}{c}{Quantized model (Baseline 1) }\\ \hline
		16 bit & 50            & 25.68  & 89.38              & -            \\
		8 bit & 75            & 16.41  & 88.86           & -             \\ \hline
		\multicolumn{5}{c}{SVD: Proposed vs Offline Compression (Baseline 2)}\\ \hline
		&               &          &   Proposed   &   Baseline2  \\ 
		DAN             & 10            & \bf{33.96}  & \bf{90.15}              & 87.35     \\
		& 30            & 29.43  & 90.12         & 87.34     \\
		& 50            & 23.16  & 89.47         & 87.14       \\
		& 70            & 18.21 & 89.55         & 86.15     \\
		& 90            & \bf{13.08}  & \bf{89.23}          & 83.72  \\ \hline
	\end{tabular}}
	\caption{\small Compression and accuracy results on the Books Intent dataset. R(\%) refers to percentage of model size reduction, Size is the model size, and Acc is the classification accuracy. All DAN models use the DAN-RAND variant.}
	\label{results:intent}
\end{table}

\subsection{Results on SST2 }
Table~\ref{results:sst} shows compression and accuracy results on the SST2 test dataset
for our proposed methods, the two baselines described in Section
~\ref{subsec:method_baselines}, and for two types of models: LSTM~(upper table
rows) and DAN-RAND~(lower table rows). For both types of models, we first report the
original uncompressed model size and accuracy.  For the quantization baseline
(baseline 1), we report size and accuracy numbers for 8-bit and 16-bit
quantization. For the SVD-based methods, both for the offline embedding
compression (baseline 2) and our proposed compression method, we compare size
and accuracy for different percentages of model size reduction $R$, where $R=1-p$. This reduction
corresponds to different fractions $p$ and to different number of selected singular values
$k$ (Section \ref{sec:method_svd}).

%Specifically, for LSTM the uncompressed model achieves $86.03$ accuracy which is state-of-the-art for this task and this model \cite{tai2015improved}. We can make similar observations for the DAN-RAND model, where the uncompressed model achieves $84.61$ accuracy which is higher than state of the art for this task and this model (\cite{iyyer2015deep} used DAN-RAND on SST2 and achieved $83.2$). 

From Table~\ref{results:sst}, we observe that our proposed compression method
outperforms both baselines for both types of models. For LSTM, our method achieves
$R=90\%$ compression with $85.11$ accuracy e.g., only 1\% rel. degradation
compared to the uncompressed model, while the offline embedding compression
(baseline 2) achieves $82.54$ (4\% rel degradation vs uncompressed). Looking at
$R=50\%$ our method achieves accuracy of $85.67$ (0.4\% rel degradation) vs
$84.24$ for baseline 2 (2\% rel degradation) and $85.08$ for baseline 1 which
does 16-bit compression (1\% rel degradation).  For DAN-RAND and for $R=90\%$ compression, our method achieves accuracy of $83.11$ which corresponds to 1.7\% rel
degradation vs an uncompressed model, and outperforms baseline 2 that has
accuracy of $82.59$ (2.4\% rel. degradation). Similar trends are seen across
various size reduction percentages $R$. \\
Comparing LSTM vs DAN-RAND models in terms of the gain we achieve over quantization, we observe that for LSTM our compression outperforms quantization by a large margin for the same compression rates, while for DAN-RAND the gain is less pronounced. We hypothesize that this is because LSTMs have recurrent connections and a larger number of low precision multiplications compared to DAN, which has been shown to lead to worse performance\cite{lin2015neural}. Thus, we expect our method to benefit the most over quantization when the network is deep.

% \begin{figure*}[t]
%        \centering
%         \begin{subfigure}[b]{0.45\textwidth}
%                 \centering
%                 \includegraphics[width=\linewidth]{alexa_lstm.png}
%                 \caption{LSTM model}
%                 \label{fig:lstm_alexa}
%         \end{subfigure}%
%         \begin{subfigure}[b]{0.45\textwidth}
%                 \centering
%                 \includegraphics[width=\linewidth]{alexa_lstm.png}
%                 \caption{DAN model}
%                 \label{fig:dan_alexa}
%         \end{subfigure}%
%         \caption{Dev set accuracy vs training mini batches for various compression percentages R for the Books Intent dataset.}
%         \label{fig:alexa_plots}
% \end{figure*}
\begin{figure*}[]
	\centering
	\begin{subfigure}[b]{0.45\textwidth}
		\centering
		\includegraphics[width=\linewidth]{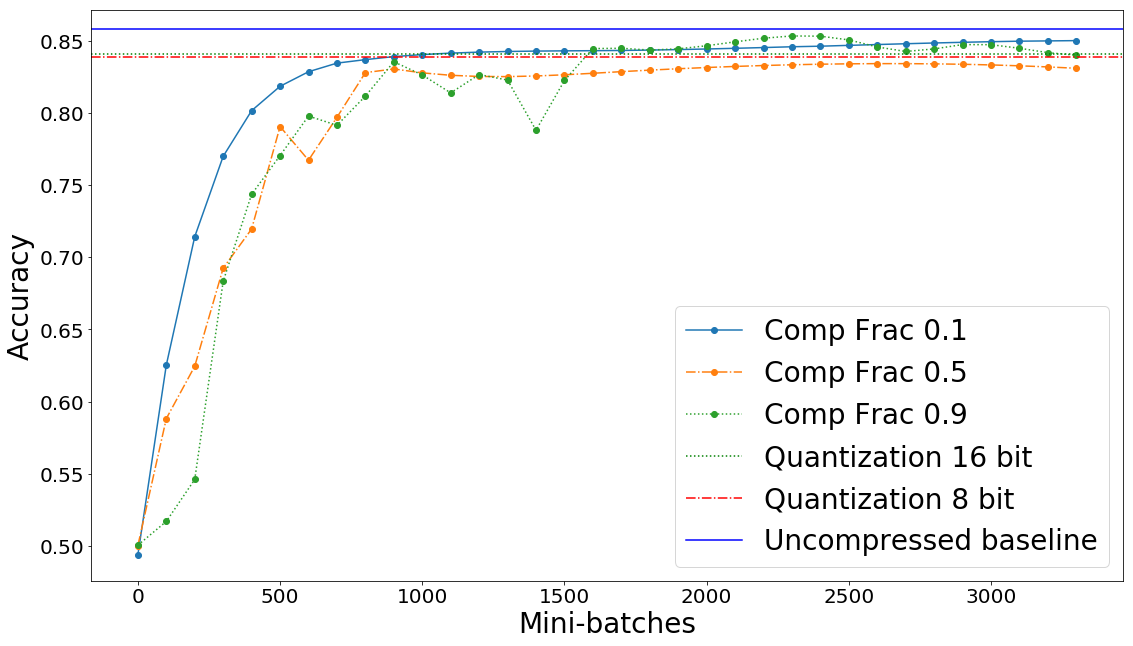}
		\caption{LSTM model}
		\label{fig:lstm_sst2}
	\end{subfigure}%
	\begin{subfigure}[b]{0.45\textwidth}
		\centering
		\includegraphics[width=\linewidth]{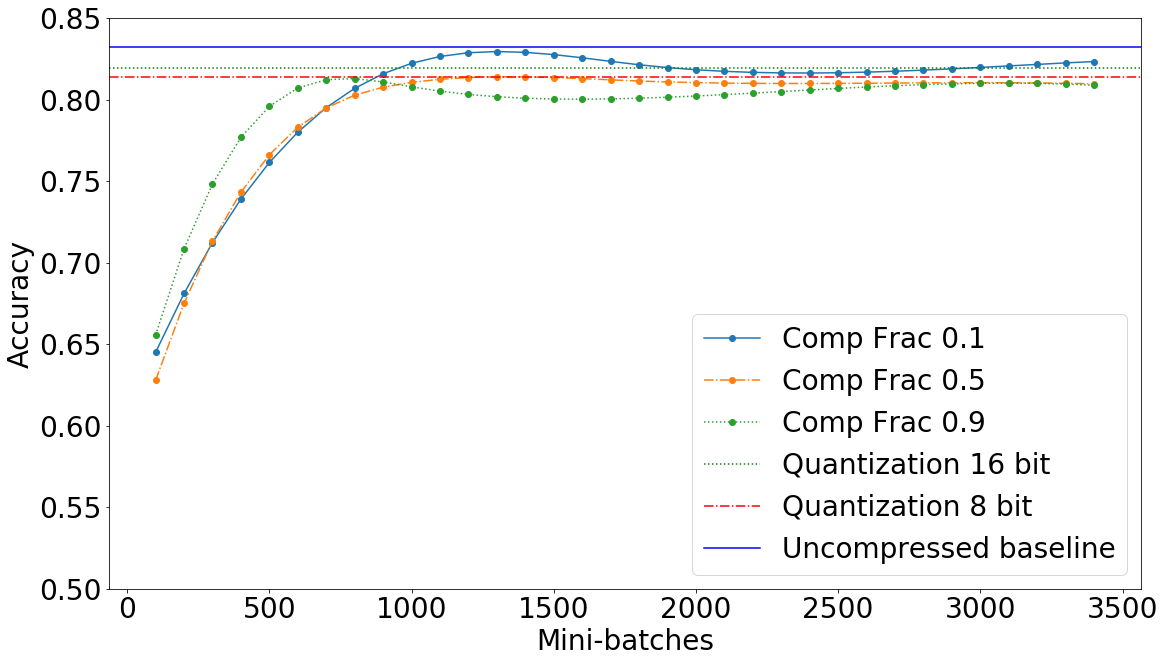}
		\caption{DAN-RAND model}
		\label{fig:dan_sst2}
	\end{subfigure}%
	\caption{\small Dev set accuracy vs training mini batches for various compression percentages R for the SST2 dataset}
	\label{fig:sst2_plots}
\end{figure*}

To illustrate how our proposed method benefits from re-training the model after
compressing the embedding input matrix, in Fig~\ref{fig:sst2_plots} we plot the
dev set accuracy for SST2 across training mini-batches for different compression
percentages R and for both LSTM and DAN-RAND. For comparison, we plot the dev
accuracy of 8-bit and 16-bit quantization (baseline 1) that remains stable
across epochs as the quantized model is not retrained after compression. As a
further comparison, we plot the uncompressed model accuracy on the dev which is
kept stable as well (we assume this as our benchmark point). We observe that,
while our compressed model starts off at a lower accuracy compared to both the
uncompressed model and the quantization baseline, it rapidly improves with
training and stabilizes at an accuracy that is higher than the quantization
method and comparable to the original uncompressed model. This indicates that
re-training after SVD compression enables us to fine-tune the layer parameters
$W_a$ and $W_b$ towards the target task, and allows us to recover most of the
accuracy.

\vspace{-1mm}
\subsection{Results on Books Intent Dataset}
Table~\ref{results:intent} shows compression and accuracy results on the
proprietary Books Intent test set for our proposed methods vs the two baselines,
for the LSTM and DAN-RAND models. Overall, we make similar observations as for
the SST2 dataset. Our proposed method achieves better accuracy across
compression percentages compared to both offline embedding compression and
quantization for both LSTM and DAN-RAND. For example, for LSTM and R=90\% compression we
achieve accuracy of 90.94\% which corresponds to 1\% degradation compared to the
uncompressed model, while offline embedding compression (baseline 2) achieves
accuracy of 89.83 (2\% degradation). We also examined the plots of dev set
accuracy while re-training the proposed compressed model across mini-batches and
observed similar trends as for SST2 (plots are omitted for brevity).

\vspace{-1mm}
\subsection{Results on Inference Time}
%We don't include the development accuracy tuning plots but we observe similar trends as the SST dataset. 
In Table~\ref{results:time_analysis}, for both LSTM and DAN-RAND, we report inference times for the SST2 test set for different compression fractions using the proposed compression vs inference times for 16-bit and 8-bit fixed point quantization~(baseline 2). As expected, for our method we see a decrease in inference time as the compression fraction~(R) increases. We observe similar trends for the quantization method, where 8-bit is faster
than 16-bit. For similar compression rates~(16-bit equivalent to 50\%
compression and 8-bit is equivalent to 75\% compression) our method and baseline 2 (fixed-point-quantization) show similar inference times. Therefore, our method does not introduce any significant latency during inference while regaining the accuracy. 
\begin{table}[]
  \centering
  \scalebox{0.9}{
	\begin{tabular}{lllll} \hline
		& LSTM && DAN-RAND \\ \hline
		R(\%) & Proposed & Baseline 2 &  Proposed & Baseline 2\\ \hline 
		10 &  10.71 & -. & 1.38 & -  \\
		50 & 9.28 & 9.23 & 0.89 & 0.93\\
		70 & 9.21 & -. & 0.62 & -\\
		75 & 9.11 &  9.08 & 0.59 & 0.62\\
		90 & 8.46 & - & 0.45 & -\\		\hline	
	\end{tabular}	}
	\caption{\small Inference Time on the SST2 test set (seconds)}
	\label{results:time_analysis}
\end{table}

%% file: conclusion.tex
\section{Conclusions and Future Work}

In this work, we have proposed a neural model compression method based on low
rank matrix factorization that can reduce the model memory footprint by an
arbitrary proportion, that can be decided based on accuracy-memory
trade-offs. Our method consists of compressing the model and then re-training it
to recover accuracy while maintaining the reduced size. We have evaluated this
approach on text classification tasks and showed that we can achieve up to 90\%
model size compression for both LSTM and DAN models, with minimal accuracy degradation (1\%-2\% relative) compared to an uncompressed model. We also showed that our
method empirically outperforms common model compression baselines such as
fixed point quantization and offline word embedding compression for the classification
problems we examined. We have also provided an analysis of our method's
effect on the model inference time and showed under which conditions it can
achieve faster inference compared to model quantization techniques. An additional contribution of this work is the introduction of a novel learning rate scheduling algorithm, the Cyclically Annealed Learning Rate (CALR). We compared CALR to other popular adaptive learning rate algorithms and showed that it leads to better performance on the SST2 benchmark dataset. In future,
we plan to evaluate the proposed compression method and the proposed CALR schedule on a larger variety of NLP tasks including sequence tagging and sequence to sequence modeling.